%% file: IGSC2020.tex
\documentclass[conference]{IEEEtran}
\ifCLASSINFOpdf
\else
\fi
\usepackage{amsmath,amssymb,amsfonts}
\usepackage[caption=false]{subfig}
\usepackage{threeparttable}
\usepackage{multirow}
\usepackage{cite}
\usepackage{microtype}
\usepackage{dblfloatfix}
\usepackage{tikz}

\usepackage{pifont}
\let\oldding\ding
\renewcommand{\ding}[2][1]{\scalebox{#1}{\oldding{#2}}}
\newcommand{\ineq}[1]{\footnotesize$#1$\normalsize}{}

{}
\renewcommand{\baselinestretch}{0.91}

\begin{document}
%
\title{Compiling Spiking Neural Networks to Mitigate Neuromorphic Hardware Constraints}

\author{\IEEEauthorblockN{Adarsha Balaji}
\IEEEauthorblockA{Department of Electrical and Computer Engineering\\
Drexel University\\
Philadelphia, Pennsylvania, USA\\
adarsha.balaji@drexel.edu}
\and
\IEEEauthorblockN{Anup Das}
\IEEEauthorblockA{Department of Electrical and Computer Engineering\\
Drexel University\\
Philadelphia, Pennsylvania, USA\\
anup.das@drexel.edu}
}


%


\IEEEoverridecommandlockouts
\IEEEpubid{978-0-7381-4307-1/20/\$31.00 ©2020 IEEE}

\maketitle

\begin{abstract}
\input{sections/abstact}
\end{abstract}


%
\IEEEpeerreviewmaketitle

\section{Introduction}\label{sec:introduction}
\input{sections/introduction}

\section{Methodology and Results}\label{sec:results}
\input{sections/results}

\ifCLASSOPTIONcaptionsoff
  \newpage
\fi

\bibliographystyle{IEEEtran}
\bibliography{commands,disco,external,snnhw,sch_anup}

\end{document}

%% file: sections/abstact.tex
Spiking Neural Networks (SNNs) are efficient computation models to perform spatio-temporal pattern recognition on {resource}- and {power}-constrained platforms. SNNs executed on neuromorphic hardware can further reduce energy consumption of these platforms. 
With increasing model size and complexity, mapping SNN-based applications to tile-based neuromorphic hardware is becoming increasingly challenging. This is attributed to the limitations of neuro-synaptic cores, viz. a crossbar, to accommodate only a fixed number of pre-synaptic connections per post-synaptic neuron. For complex SNN-based models that have many neurons and pre-synaptic connections per neuron,\\ (1) connections may need to be pruned after training to fit onto the crossbar resources, leading to a loss in model quality, e.g., accuracy, and (2) the neurons and synapses need to be partitioned and placed on the neuro-sypatic cores of the hardware, which could lead to increased latency and energy consumption. 
In this work, we propose (1) a novel unrolling technique that decomposes a neuron function with many pre-synaptic connections into a sequence of homogeneous neural units to significantly improve the crossbar utilization and retain all pre-synaptic connections, and (2) SpiNeMap, a novel methodology to map SNNs on neuromorphic hardware with an aim to minimize energy consumption and spike latency.

%% file: sections/introduction.tex
\IEEEPARstart{S}{piking} Neural Networks (SNNs) are machine learning approaches designed using spike-based communication and computation, and bio-inspired learning algorithms~\cite{maass1997networks}. Neurons are typically implemented using Integrate-and-Fire \cite{chicca2003vlsi} or Izhikevich \cite{izhikevich2003simple} models. The neurons are interconnected using weighted synapses and communicate by sending short impulses, called spikes, over the synapse. Supervised, semi-supervised, or an unsupervised approach \cite{kasabov2001evolving,lee2016training,mostafa2018supervised} can be employed to train the SNN. Information in SNNs can be represented as inter-spike interval or spike rates \cite{rullen2001rate,dorval2008probability,peper2016low}. SNNs can be trained and inferred on the general purpose hardware \cite{pycarl}, such as CPUs and GPUs. However, due to the \emph{high} resource and energy demands, the use of this type of hardware in constrained environments is often \emph{prohibitive}.

Neuromorphic hardware \cite{chicca2014neuromorphic} (1) reduces the energy consumption significantly, due to their \emph{low-power} design of neurons and synapses, and (2) improves application throughput, due to their \textit{distributed} implementation of computation and storage. In recent years, several neuromorphic hardware platforms are designed: DYNAP-SE \cite{Moradi_etal18}, TrueNorth \cite{akopyan2015truenorth}, and Loihi \cite{davies2018loihi}. 
Although these hardware differ in their operation (e.g., analog vs. digital), they all support neuro-synaptic cores realised using crossbar-based architectures. Multiple crossbars are interconnected over a time-multiplexed interconnect \cite{dynapse, balajiexploring2019}.

A crossbar is a two-dimensional arrangement of synapses ($n^2$ synapses for $n$ neurons). At the cross-point in the crossbar there is a synaptic element, which can be implemented using nano-scale Non-Volatile Memory (NVM)~\cite{Mallik2017}. Within each crossbar, synaptic weights between pre- and post-synaptic neurons are programmed as conductance of NVM cells. Scaling up the size of a crossbar increases the number of synapses ($n$) per neuron, which exponentially increases the dynamic and leakage energy. 
To keep energy consumption reasonable, neuromorphic engineers limit the size of a crossbar. For instance, in TrueNorth and DYNAP-SE, the crossbar can accommodate a fixed number of synapses per neuron. To build large neuromorphic hardware, the common practice is therefore to integrate multiple crossbars, with the shared, time-multiplexed interconnect. 
While executing SNNs on crossbar-based neuromorphic hardware we encounter the following constraints: 
\vspace{0.2mm} \\

\textit{Constraint 1: Crossbars on Neuro-synaptic cores can only accommodate a limited number of synaptic connections per post-synaptic neuron.}

A \ineq{n\times n} crossbar in a tile can accommodate only \ineq{n} pre-synaptic connections per post-synaptic neuron (\ineq{n = 128} for the crossbars in DYNAP-SE).
Model pruning is proposed as a solution. However, we observe that, even after model pruning, on average {2.6\%} of neurons in these models cannot be mapped to the crossbars in DYNAP-SE. There are two currently-used solutions to this problem -- 1) implement larger crossbars, which increases the power consumption exponentially, and 2) remove synaptic connections, which reduces the model quality.

\textit{Constraint 2: The neurons and synapses of the SNN need to be efficiently partitioned and mapped to resources of the neuromorphic hardware.}

Typically, the number of neurons and synapses of realistic SNN applications \cite{HeartClassJolpe, Das2018HeartbeatCI, das2018unsupervised} far exceed the  resources available on a single neuro-synaptic core (crossbar). Therefore, to execute this application on the neuromorphic hardware, the SNN first needs to be partitioned into clusters of neurons and synapses, where the intra-cluster \emph{local synapses} are mapped within the crossbars, and the inter-cluster \emph{global synapses} are mapped on the shared interconnect. An inefficient placement (ad-hoc) of the SNN on the crossbars of the hardware can increase energy consumption, and reduce application accuracy by increasing spike latency.

 \newpage
In this thesis, we address the constraints by:

\begin{itemize}
    \item A novel unrolling technique \cite{balaji2020enabling} to decompose a neuron function with many pre-synaptic connections into a sequence of homogeneous neural units, where each neural unit is a function computation node, with two pre-synaptic connections.
    \item \textbf{SpiNeMap} \cite{balaji2019mapping}: an end-to-end methodology to partition an SNN into clusters of local and global synapses, and place the clusters on the crossbars, with an aim to minimize the energy consumption and increase the application accuracy.  
\end{itemize}



%% file: sections/results.tex
\subsection{Neuron Spatial Decomposition Methodology}
We propose an {unrolling} approach \cite{balaji2020enabling}, which decomposes a neuron function computation with many fan-ins into a sequence of homogeneous \textit{neural units}, where each neural unit is a computation node with a maximum fanin-of-two ({FIT}). Here, one \ineq{m}-input neuron function is decomposed into \ineq{(m-1)} two-input neural units connected in sequence.

The neuron function 

\begin{equation}
    \label{eq:original_nn}
    \footnotesize y_o = \varphi\left(\sum_{i=1}^m n_i\cdot w_i\right)
\end{equation}

is represented as

\begin{equation}
\label{eq:decomposed_nn}
\footnotesize y_o = f(u_{m-1}), \\ 
        \text{ where } \\  u_i = \begin{cases}
        f(n_1\cdot w_1 + n_2\cdot w_2), \text{ for } i = 1
        \\
        f(u_{i-1} + n_{i+1}\cdot w_{i+1}), \text{ otherwise}.
        \end{cases}
\end{equation}

where, \ineq{f} represents the neuron functionality of generating spike trains with a mean firing rate proportional to its input excitation, \ineq{n_1,n_2,\cdots,n_m} are the \ineq{m} pre-synaptic neurons of the post-synaptic neuron \ineq{n_o}, and \ineq{w_1,w_2,\cdots,w_m} are the corresponding synaptic weights. The total number of FIT neural units generated from a neural network with \ineq{N} neurons is

\begin{equation}
    \label{eq:decomposed_nn1}
    \footnotesize \mathcal{N} = \sum_{i=1}^N(m_i-1), \\ \text{ where } \\ m_i \text{ is the fan-in of neuron } \\ n_i
\end{equation}

\subsection{Spatial Decomposition Results}
The unrolling technique significantly improves crossbar utilization  and  ensures  information  integrity,  resulting  in  no loss in model quality derived from connection pruning. We integrate  this  unrolling  technique  inside  an  existing  SNN mapping  framework  and  evaluate  it  using  machine  learning applications for a state-of-the-art neuromorphic hardware. Our results demonstrate an average 60\% lower crossbar requirement, 9x higher synapse utilization, 62\% lower wasted energy, and 3\% increase in model quality compared to an existing SNN mapping approach.

\subsection{SpiNeMap Methodology}
The SpiNeMap methodology partitions SNNs into clusters of local synapses, and place these clusters on the physical crossbars such that energy consumption is minimized and application accuracy is increased. The contributions of SpiNeMap are as follows:

\begin{itemize}
	\item \textbf{SpiNeCluster}: We propose a greedy heuristic to partition SNNs into local and global synapses with the objective of reducing the total number of global spikes. 
	
	We represent the SNN as a set, G(N,S), of \textit{N} neurons and S synapses. We are interested in partitioning this SNN into $k$ clusters, such that the total number of inter-cluster spikes is minimized. The partitioning problem is treated as a graph partitioning problem \cite{kernighan1970efficient}, and has been applied in many context, including task mapping on multiprocessor systems \cite{das2014communication}. The graph partitioning problem is already NP-complete \cite{garey1974some}, so heuristics are typically used to solve them \cite{fiduccia1982linear}. We propose a greedy approach, roughly based on the Kernighan-Lin Graph Partitioning algorithm \cite{kernighan1970efficient}, which we show to be scalable to large SNNs.
	\item \textbf{SpiNePlacer}: We propose  a PSO-based cluster placement approach to place the local synapses inside the physical crossbars, and time-multiplex the global synapses on the shared interconnect. The objective is to reduce energy consumption, spike latency, and spike disorder.
	
	PSO finds the optimum solution to a fitness function $F$. Each solution is represented as a particle in the swarm. Each particle has a velocity with which it moves in the search space to find the optimum solution. During the movement, a particle updates its position and velocity according to its own experience (closeness to the optimum) and also experience of its neighbors. 
	
	The objective function $F$ for SpiNePlacer, which is to minimize (1) energy consumption ($E$) on global synapses (i.e., the interconnect), (2) ISI distortion ($I$), and (3) spike disorder ($D$), for a given placement of the clusters in a partitioned SNN. We note that reducing the ISI distortion and spike disorder increases the application accuracy. We express SpiNePlacer's \textbf{objective function} as

    \begin{equation}
    \label{eq:pso_obj_spine}
    \footnotesize F(\mathcal{P}^t_{\mathcal{H}}) = E\times I\times D,
    \end{equation}

    where $\mathcal{P}^t_{\mathcal{H}}$ is the placement of the partitioned SNN $\mathcal{H(C,E)}$ at the $t^{\text{th}}$ iteration of the PSO.
	
\end{itemize}

\subsection{SpiNeMap Results}
We show that SpiNeMap reduces energy consumption by 51\% and spike latency by 17\%, compared to the state-of-the-art techniques. Overall, SpiNeMap improves application accuracy between  1\% and 26\% (average 10\%) for all the evaluated SNNs, executed on the DYNAP-SE platform.

\section{Conclusion}
In this thesis we propose a compiler framework for SNN-based application executed on tile-based neuromorphic hardware. The compiler performs a novel unrolling technique that decomposes a SNN neuron function with many pre-synaptic  connections  into  a  sequence  of  homogeneous neural units to significantly improve the crossbar utilization and retain  all  pre-synaptic  connections followed by a  novel methodology  to  map the neurons and synapses of the SNN  on  neuromorphic  hardware  with  an aim  to  minimize  energy  consumption  and  spike  latency. The proposed SNN compiler can be integrated with many other mapping approaches such as the data flow-based SNN compilation technique of \cite{song2020compiling, balaji2020run, balaji2019framework, das2018dataflow, das2018mapping}, the circuit aging oriented SNN mapping technique of
\cite{twisha2020reliability, song2020improving, twisha2020thermal, balaji2019framework2, song2020dependability}, and the SNN compiler of \cite{ji2018bridge}.